\documentclass[10pt,twocolumn,letterpaper]{article}

\usepackage[pagenumbers]{iccv}              

\usepackage{multirow}
\usepackage{makecell}

\definecolor{iccvblue}{rgb}{0.21,0.49,0.74}
\usepackage[pagebackref,breaklinks,colorlinks,allcolors=iccvblue]{hyperref}

\title{Efficient Long-Tail Learning in Latent Space by sampling Synthetic Data}

\author{Nakul Sharma\\
Independent Researcher\\
{\tt\small 0xnakul.sh@gmail.com}
}

\begin{document}
\maketitle
\begin{abstract}
Imbalanced classification datasets pose significant challenges in machine learning, often leading to biased models that perform poorly on underrepresented classes. With the rise of foundation models, recent research has focused on the full, partial, and parameter-efficient fine-tuning of these models to deal with long-tail classification. Despite the impressive performance of these works on the benchmark datasets, they still fail to close the gap with the networks trained using the balanced datasets and still require substantial computational resources, even for relatively smaller datasets. Underscoring the importance of computational efficiency and simplicity, in this work we propose a novel framework that leverages the rich semantic latent space of Vision Foundation Models to generate synthetic data and train a simple linear classifier using a mixture of real and synthetic data for long-tail classification. The computational efficiency gain arises from the number of trainable parameters that are reduced to just the number of parameters in the linear model. 
Our method sets a new state-of-the-art for the CIFAR-100-LT benchmark and demonstrates strong performance on the Places-LT benchmark, highlighting the effectiveness and adaptability of our simple and effective approach.
\end{abstract}    
\section{Introduction}
\label{sec:intro}

Long-tail classification addresses the reality that real-world data often follow highly skewed distributions: a few head classes have abundant samples, whereas many tail classes are represented by only a handful of examples. The importance of long-tail learning cannot be overstated, as models trained on such imbalanced datasets often exhibit biased performance, excelling on head classes while struggling with tail classes. This bias can lead to severe consequences in critical applications such as medical diagnosis, autonomous driving, and financial fraud detection, where accurate prediction across all classes is paramount. Despite extensive research spanning data re-balancing (resampling or augmentation) strategies, improved representation learning, and adjusted loss functions, the tail-class performance still lags far behind what is achieved on balanced data.

The field of long-tail learning has witnessed significant advancements in recent years, evolving from traditional approaches like data resampling and loss reweighting to more sophisticated techniques leveraging the power of deep learning.
Recently, large pre-trained vision models like CLIP~\cite{radford2021CLIP} have emerged as powerful visual backbones for long-tail recognition. These models are trained on massive, diverse datasets and produce high-quality feature embeddings, which can be leveraged to mitigate class imbalance. By fine-tuning a foundation model (instead of training from scratch on an imbalanced set), researchers have reported substantial gains in overall accuracy~\cite{ma2021BALLAD, wang2024DECODER, shi2024LIFT, shi2025lift+}. However, the challenge is how to adapt these models without undoing their benefits for tail classes. Recent studies have revealed that heavy fine-tuning of CLIP-based ViT~\cite{dosovitskiy2021ViT} model on imbalanced data can distort the feature space and actually degrade tail-class accuracy~\cite{shi2024LIFT} -- to avoid this pitfall, the authors advocate for lightweight fine-tuning. 

In this work, we look at the problem of long-tail learning from the perspective of generating novel data samples for minority classes to balance the training set, and to effectively utilize the latent space of foundation models like CLIP~\cite{cherti2023openclip}. To this end, we  propose a novel long-tail learning approach that operates entirely in the latent feature space of a frozen vision foundation model. Specifically, we use a pre-trained CLIP ViT encoder to embed all images into a high-dimensional feature space rich in semantic information. Rather than fine-tuning the backbone on the imbalanced data, we keep the feature extractor fixed and address the imbalance by generating synthetic features for all the classes so that their cardinality becomes equal. For each minority class, we perform a simple kernel density estimation (KDE) on its few available feature vectors, using von Mises-Fisher kernels to account for the directional nature of the normalized CLIP embeddings. This KDE effectively models a smooth manifold of plausible feature vectors around the known samples. We then sample additional feature points from this estimated distribution, yielding ``synthetic data'' for the tail class without ever generating images. Finally, we train a linear classifier on the augmented dataset composed of the original real features plus the synthetic features for the tail classes. The classifier training is a lightweight optimization, and by construction, it sees a much more balanced training set.

Our novelty lies in generating synthetic training examples in the latent space of a powerful vision model and using an extremely simple learning algorithm (linear classification) to achieve high accuracy on the CIFAR-100-LT and the Places-LT benchmark. This stands in contrast to prior works that either fine-tune large networks or train complex generative models to augment data for class-balancing. By avoiding any heavy network updates and repeated forward passes, our approach is efficient in computation and memory. It requires neither external data nor multi-stage training; in fact, once features are extracted (which can be done in a single forward pass per image), the rest of the training pipeline is akin to training a logistic regression on an augmented feature set. Furthermore, our approach is model-agnostic, allowing it to be seamlessly integrated with future vision foundation models and to deliver ongoing, incremental gains across a wide range of imbalanced datasets. To summarize, our contributions are as follows: i) We introduce a simple approach to efficiently utilize the pre-trained embedding space of vision foundation models, ii) We empirically validate our approach on the CIFAR-100-LT and Places-LT benchmarks, achieving state-of-the-art and competitive performance, respectively, and iii) We ablate the choice of encoders and the latent sampling to quantify the components that make our approach work better.

\section{Methodology}
\label{sec:method}
Our method leverages the pre-trained vision encoder of OpenCLIP~\cite{cherti2023openclip} to obtain the latent embeddings of all images. Since empirical evidence suggests that  normalized visual embeddings for similar images obtained from this contrastive model lie close to each other on a unit hypersphere~\cite{DreamSim, mayilvahanan2024doesCLIPSim}, we estimate the latent distribution of each class using a mixture of von Mises-Fisher (vMF) distribution kernels. The vMF distribution is one of the simplest parametric distributions for hyperspherical data, and has properties analogous to those of the multi-variate Gaussian distribution for data in $\mathbb{R}^d$~\cite{rao-stats-apps, mardia}. Once estimated, we sample new data points in latent space from the estimated distribution such that the number of samples in all classes of the data becomes equal. These newly sampled latents serve as novel synthetic data for each class, and are used to train a logistic regression model along with the original latents for classification. The following sections detail the extraction of latent embeddings, estimation of the latent distribution for each class, and generation of synthetic data.\\

\noindent \textbf{Latent Embedding Extraction}. For a given dataset with $|C|$ classes, let $I^k$ denote the set of images that belong to class $k$, where $k \in \{0, 1, \dots ,C-1\}$.
For each image $\mathbf x \in I^{k}$, we compute its latent embedding $\mathbf z \in \mathbb{R}^d$ using the frozen vision encoder $f_{\theta}$ and $\ell_{2}$--normalize it:
\begin{equation}
\mathbf z = \frac{f_{\theta}(\mathbf x)}{\lVert f_{\theta}(\mathbf x) \rVert_{2}} \in \mathbb{S}^{d-1}, \quad 
\mathbf z \in Z^{k} = \bigl\{ \mathbf z_{i}^{k} \bigr\}_{i=1}^{N_{k}}.
\label{eq:latent}
\end{equation}
Here $\mathbb{S}^{d-1}$ is the unit hypersphere in $\mathbb{R}^{d}$ and $N_{k}=|I^{k}|$. \\

\noindent \textbf{Class--Conditional Density Estimation}. 
Since all embeddings lie on $\mathbb{S}^{d-1}$, we approximate the density of the latent embeddings of class~$k$ using the Kernel Density Estimation~\cite{KDE} with kernels given by von Mises-Fisher distributions centered at each of the observed latent embeddings:
\begin{equation}
\hat{p}_{k}(\boldsymbol{\cdot})
= \frac{1}{N_{k}} \sum_{i=1}^{N_{k}} \mathrm{vMF}\bigl(\mathbf z; \boldsymbol \mu = \mathbf z_{i}^{k}, \kappa = \kappa_{\mathbf z_i^{k}}\bigr),
\label{eq:kde}
\end{equation}
where the von Mises-Fisher (vMF) probability density function is defined as:
\begin{equation}
\mathrm{vMF}(\mathbf z; \boldsymbol{\mu}, \kappa)
= C_{d}(\kappa) \exp\bigl(\kappa \boldsymbol {\mu}^{\top} \mathbf z\bigr),
\label{eq:vmf1}
\end{equation}
where $
\quad
C_{d}(\kappa)
= \frac{\kappa^{d/2-1}}{(2\pi)^{d/2} I_{d/2-1}(\kappa)},
\label{eq:vmf2}
$
 and $\kappa > 0$ controls the concentration of the distribution, and $I_{\nu}(\cdot)$ denotes the modified Bessel function of order $\nu$. To estimate vMF KDE in Equation~\eqref{eq:kde}, we need to estimate the concentration $\kappa_{\mathbf z^k_i}$ at each embedding $\mathbf z_i^k$. \\

\noindent \textbf{Estimating the Concentration $\kappa_{\mathbf z^k_i}$}.
~\citet{JMLR:v6:banerjee05clustering} derives that the concentration $\kappa_{k}$ for the observations $Z^k$ can be approximated as
$\kappa_{k} \approx \frac{\bar{R}_{k}(d - \bar{R}_{k}^{2})}{1 - \bar{R}_{k}^{2}}
\label{eq:kappa}
$
where $\bar{R}_{k}$ is the sample resultant length given by
$
\bar{R}_{k}=\Bigl\lVert \frac{1}{N_{k}}\sum_{i=1}^{N_{k}} \mathbf z_{i}^{k}\Bigr\rVert_{2}.
$ This closed-form solution for $\kappa_k$ yields accurate estimates with an almost negligible computational cost. However, we are interested in estimating $\kappa_{\mathbf z^k_i}$ in Equation~\eqref{eq:kde} which is entirely different from $\kappa_k$, the concentration estimate for the whole distribution of the $k$-th class. 

To estimate the local concentration parameter $\kappa_{\mathbf z^k_i}$ at each latent embedding $\mathbf z_i^{k}$, we exploit the empirical observation that latent embeddings of visually similar images exist closely on the unit hypersphere. Specifically, for each embedding $\mathbf z_i^{k}$, we first identify its nearest neighbor $\mathbf z_j^{k}$ from the same class $k$ in the latent space using cosine similarity such that
$\mathbf z_j^{k} = \arg\max_{\mathbf z \in \mathbf Z^{k} \setminus {\mathbf z_i^{k}}} (\mathbf z_i^k)^{\top} \mathbf z.
\label{eq:nearest}
$

Since these embeddings belong to same class $k$ and $\mathbf z_i^{k}$ is closest to $\mathbf z_j^{k}$, we assume that both of these embeddings originate from the same underlying concentrated latent vMF distribution. Under this assumption, we form the local dataset $\tilde{Z}_i^{k} = \{ \mathbf z_i^{k}, \mathbf z_j^{k} \}$ and estimate the concentration parameter $\kappa_{\mathbf z_i^{k}}$ using the method proposed by Banerjee et al.~\cite{JMLR:v6:banerjee05clustering} as follows:
\begin{equation}
\kappa_{\mathbf z_i^{k}} \approx \frac{\tilde{R}_{i}^{k}(d - (\tilde{R}_{i}^{k})^{2})}{1 - (\tilde{R}_{i}^{k})^{2}}, \quad \text{where} \quad \tilde{R}_{i}^{k} = \Bigl\lVert \frac{1}{2}(\mathbf z_i^{k} + \mathbf z_j^{k}) \Bigr\rVert_2.
\label{eq:kappa_local}
\end{equation}

\noindent The calculated $\kappa_{\mathbf z_i^{k}}$ thus reflects the local concentration at the embedding $\mathbf z_i^{k}$, and helps in effectively capturing variations in density across different regions of the class-conditional latent space. These locally estimated concentrations are then used in Equation~\eqref{eq:kde} to accurately model the class-specific distribution of latent embeddings. \\

\noindent \textbf{Synthetic Latent Generation}
After estimating the density using Equation \eqref{eq:kde}, we generate synthetic latent embeddings to balance each class. Specifically, we draw $\tilde{N}_{k} = N_{\max} - N_{k}$ synthetic samples per class, where $N_{\max} = \max_{c} N_{c}$, ensuring all the classes reach equal cardinality. We utilize Wood’s rejection sampling method~\cite{Wood1994SimulationOT-rejsample} to efficiently sample from the estimated mixture of vMF distributions for each class $k$. The augmented and balanced latent set $\mathcal{Z}^{k}$ for class k is expressed as follows:
\begin{equation}
\mathcal{Z}^{k}=Z^{k} \cup \bigl\{\tilde{\mathbf z}_{j}^{k}\bigr\}_{j=1}^{\tilde{N}_{k}}, 
~~~~\tilde{\mathbf z}_{j}^{k} \sim \hat{p}_{k}(\boldsymbol{\cdot}),
\label{eq:balanced}
\end{equation}


\noindent where $\hat{p}_{k}(\boldsymbol{\cdot})$ is defined in Equation~\eqref{eq:kde}. After generating synthetic samples for each minority class, we merge the balanced embedding sets into $\mathcal{Z}$ and fit a multinomial logistic-regression classifier $W_\phi$ with L–BFGS. At inference, we encode an image $I_t$, $\ell_2$-normalize its embedding $z = f_\theta(\mathbf I_t)/\| f_\theta(\mathbf I_t) \|_2$, and predict $\hat{\mathbf y} = \arg\max W_\phi\left(\mathbf z\right)$.


\section{Experiments}
\label{sec:experiments}

We conduct comprehensive experiments to demonstrate the effectiveness, robustness, and versatility of our proposed method. Our evaluation is focused on widely used long-tail benchmarks, specifically CIFAR-100-LT~\cite{cao2019learning-cifar100lt} and Places-LT~\cite{openlongtailrecognition-placeslt}, chosen due to their prevalence and challenging nature in the domain of imbalanced classification tasks. Performance is primarily assessed using top-1 accuracy, offering a clear and direct measure of classification effectiveness. Additionally, we conduct detailed ablation studies to gain deeper insights into the influence of encoders and synthetic data generation methodologies.\\

\noindent \textbf{Experimental Setup}. For extracting latent embeddings, we employ the Vision Transformer ViT-L/14 encoder from the OpenCLIP~\cite{cherti2023openclip}, due to its proven ability to capture discriminative visual features across diverse datasets. To model the latent embeddings, we adopt a mixture of vMF distributions, implementing our approach by extending the base \texttt{vonmises\_fisher} class available in the SciPy library~\cite{2020SciPy-NMeth}. For classifier training, we utilize logistic regression optimized through the L-BFGS algorithm~\cite{LBFGS}, using the scikit-learn library~\cite{scikit-learn}.\\

\begin{table}[!t]
\caption{Comparison with state-of-the-art methods on CIFAR-100-LT with various imbalance ratios.}
\label{table:comp_cifar100lt}
\setlength{\tabcolsep}{1.2ex} 
\centering
\begin{small}
\begin{tabular}{l|c| c c c }
\toprule
\multirow{2}{*}{{Method}} & 
\multirow{2}{*}{\makecell{{Learnable}\\ {Params.}}} &
\multicolumn{3}{c}{{Imbalance Ratio}} \\
& &  100 & 50 &10 \\
\midrule
\multicolumn{5}{l}{\bf Training from scratch} \\
\midrule

LDAM \citep{cao2019learning-cifar100lt} & 0.46M & 42.0 & 46.6 & 58.7 \\
BBN \citep{zhou2020bbn}  & 0.46M & 42.6 & 47.0 & 59.1 \\
DiVE \citep{he2021distilling} & 0.46M & 45.4 & 51.1 & 62.0 \\
MiSLAS \citep{zhong2021improving} & 0.46M & 47.0 & 52.3 & 63.2 \\
BS \citep{ren2020balanced} & 0.46M & 50.8 & 54.2 & 63.0 \\
PaCo \citep{cui2021parametric} & 0.46M & 52.0 & 56.0 & 64.2 \\
BCL \citep{zhu2022balanced} & 0.46M & 51.9 & 56.6 & 64.9 \\
\midrule
\multicolumn{5}{l}{\bf Fine-tuning pre-trained model} \\
\midrule

LiVT \citep{xu2023learning} & 85.80M & 58.2 & - & 69.2 \\
BALLAD \citep{ma2021BALLAD} & 149.62M & 77.8 & - & - \\
LIFT~\cite{shi2024LIFT} & 0.10M & 80.3 & 82.0 & 83.8 \\
LIFT+~\cite{shi2025lift+} & 0.10M & 81.7 & 83.1 & 84.7 \\

\midrule
\multicolumn{5}{l}{\bf Utilizing frozen pre-trained model} \\
\midrule
Ours &  0.10M & \bf 89.0 & \bf 90.3 & \bf 91.4 \\
\bottomrule
\end{tabular}
\end{small}
\end{table}

\begin{table}
\caption{Comparison with state-of-the-art methods on Places-LT.}
\label{table:comp_placeslt}
\setlength{\tabcolsep}{0.5ex} 
\centering
\begin{small}
\begin{tabular}{l|c|cccc}
\toprule
 Method &\makecell{Learnable \\ Params.} & Overall & Head & Medium & Tail\\ 
\midrule
\multicolumn{6}{l}{\bf Training from scratch (init. from ImageNet-1K backbone)} \\
\midrule
OLTR \citep{liu2019large} & 58.14M & 35.9 & 44.7 & 37.0 & 25.3 \\
cRT \citep{kang2020decoupling} & 58.14M & 36.7 & 42.0 & 37.6 & 24.9 \\
LWS \citep{kang2020decoupling} & 58.14M & 37.6 & 40.6 & 39.1 & 28.6 \\
MiSLAS \citep{zhong2021improving} & 58.14M & 40.4 & 39.6 & 43.3 & 36.1 \\
DisAlign \citep{zhang2021distribution} & 58.14M & 39.3 & 40.4 & 42.4 & 30.1 \\
ALA \citep{zhao2022adaptive} & 58.14M & 40.1 & 43.9 & 40.1 & 32.9 \\
PaCo \citep{cui2021parametric} & 58.14M & 41.2 & 36.1 & 47.9 & 35.3  \\
LiVT \citep{xu2023learning} & 85.80M & 40.8 & 48.1 & 40.6 & 27.5 \\
\midrule
\multicolumn{6}{l}{\bf Fine-tuning foundation model} \\
\midrule
BALLAD \citep{ma2021BALLAD} & 149.62M & 49.5 & 49.3 & 50.2 & 48.4 \\
Decoder \citep{wang2023exploring}& 21.26M & 46.8 & - & - & - \\
LPT \citep{dong2023lpt} & 1.01M & 50.1 & 49.3 & \bf 52.3 & 46.9 \\
LIFT~\cite{shi2024LIFT} & 0.18M & \bf 51.5 & \bf 51.3 &  52.2 & 50.5 \\
LIFT+~\cite{shi2025lift+} & 0.18M & \bf 51.5 & 50.8 & 52.0 & \bf 51.6 \\

\midrule
\multicolumn{5}{l}{\bf Utilizing frozen pre-trained model} \\
\midrule
Ours & 0.37M & 48.3 & 49.3 & 48.4 & 47.0 \\
\bottomrule
\end{tabular}
\end{small}
\end{table}
\vspace{-2pt}
\noindent \textbf{Comparison with State-of-the-Art}. We benchmark our method against recent works that address long-tail learning using vision foundation models, as well as traditional approaches known for low computational demands on CIFAR-100-LT~\cite{cao2019learning-cifar100lt} and Places-LT~\cite{openlongtailrecognition-placeslt} benchmarks. 

Results for the CIFAR-100-LT benchmark in Table~\ref{table:comp_cifar100lt} show that our approach achieves a substantial performance leap over existing methods at all imbalance ratios. Notably, under the most challenging imbalance ratio ($100$), our method reaches an impressive top-1 accuracy of $89\%$, outperforming the previous best LIFT+~\cite{shi2025lift+} baseline by 7.3\%. Even at less severe imbalance ratios ($50$ and $10$), our method maintains a superior accuracy ($90.3\%$ and $91.4\%$, respectively), demonstrating its robustness across varying levels of class imbalance.

The competitive performance of our method on the Places-LT benchmark reported in Table~\ref{table:comp_placeslt} demonstrates that our density-guided synthesis strategy scales gracefully from small-scale to large-scale long-tail settings. Despite the encouraging results, our method still trails the latest adapter-based fine-tuning techniques.
We conjecture that this shortfall stems from limitations of the frozen latent embeddings themselves --- we use pooled embeddings, which could lead to residual semantic overlap in the latent space in a physical-world dataset like Places-LT where some categories share coarse features which require more fine-grained representation. In future work, we plan to investigate this further and develop strategies that enhance embedding discriminability.

In addition to our strong predictive performance, our method distinguishes itself by its computational efficiency. Unlike fine-tuning-based approaches, which require repeated forward and backward passes through all data across the full network during each epoch, our method needs only a single forward pass over the pre-trained backbone to extract embeddings. Subsequent steps are limited to efficient density modeling and lightweight classifier training, resulting in reduced compute requirements.\\ 

\begin{table}[!t]
\centering
\caption{Comparison of training strategy for ViT on CIFAR-100-LT, trained across different imbalance ratios.}
\label{tab:encoder_ablation}
\begin{tabular}{l c  ccc}
\toprule
& & \multicolumn{3}{c}{Imbalance Ratio} \\
{Method} & Size & 10 & 50 & 100 \\
\midrule
\multirow{2}{*}{{SigLIP}} & B/16 & 80.84 & 77.52 & 75.22\\
 & L/14 & 87.65 & 85.61 & 83.96  \\
\midrule

\multirow{2}{*}{{OpenCLIP}} & B/16 & 87.67 & 85.43 & 83.49  \\
 & L/14 & \bf 91.44 & \bf 90.31 & \bf 89.05  \\
\bottomrule
\end{tabular}%
\end{table}

\vspace{-2pt}
\noindent \textbf{Ablations}. 
First, we investigate how different contrastive learning objectives affect performance on the CIFAR-100-LT benchmark. We compare softmax-normalized contrastive learning backbone OpenCLIP~\cite{cherti2023openclip} against non-softmax alternative SigLIP~\cite{zhai2023siglip}.
As shown in Table~\ref{tab:encoder_ablation}, softmax-based encoders consistently outperform their non-softmax counterparts across all imbalance ratios. This gap suggests that the softmax normalization's competitive dynamics create more compact and well-separated class manifolds in the latent space---this is particularly advantageous for our vMF-based density estimation, as they reduce inter-class overlap when generating synthetic samples. Moreover, scaling to larger encoder sizes 
 yields consistent improvements, underscoring the importance of model capacity for capturing discriminative features.

Second, we investigate the impact of synthetic data generation techniques. We use OpenCLIP ViT-L/14 encoder to compare our proposed vMF-based KDE with Gaussian KDE, SMOTE~\cite{chawla2002smote}, random oversampling, and a baseline that does not generate synthetic data. Results on the CIFAR-100-LT benchmark presented in Table~\ref{tab:data_gen_ablation} clearly indicate that our proposed vMF KDE method outperforms all other synthetic data generation strategies, validating its effectiveness in accurately modeling latent distributions on a hypersphere and generating high-quality synthetic embeddings. Notably, our proposed method achieves the greatest performance improvement on CIFAR-100 with an imbalance ratio (IR) of 100, outperforming both the baseline and other competing methods. This result further highlights the effectiveness of our approach in handling highly imbalanced datasets.

\begin{table}[!t]
\centering
\caption{Ablation study of synthetic data generation methods on CIFAR-100-LT trained across different imbalance ratios.}
\label{tab:data_gen_ablation}
\begin{tabular}{l  ccc}
\toprule
& \multicolumn{3}{c}{Imbalance Ratio} \\
{Method} & 10 & 50 & 100  \\
\midrule
No Synth. Data & 90.73 & 86.66 & 84.08  \\
OS w/ replacement & 91.24 & 90.13 & 87.32 \\
SMOTE & 91.36 & 90.17 & 88.66   \\
Gaussian KDE & 91.18 & 88.99 & 86.94  \\
vMF-KDE (Ours) & \bf 91.44 & \bf 90.31 & \bf 89.05  \\

\bottomrule

\end{tabular}%
\end{table}

\section{Conclusion}
\label{sec:conclusion}

In this preliminary work, we propose a novel and efficient approach to long-tail learning by generating synthetic samples in the latent space of powerful vision foundation models. Our method uses kernel density estimation with the vMF kernel to jointly model and synthesize high-quality synthetic latent samples for minority classes, enabling the training of a simple linear classifier on a balanced and enriched feature set. This approach achieves strong empirical results on challenging long-tail benchmarks such as CIFAR-100-LT and Places-LT, while significantly reducing computational overhead compared to existing state-of-the-art methods. 
Building on our promising results, future work will focus on (i) reducing the latent semantic overlap in the pre-trained representations to learn better classifiers, and (ii) systematically evaluating and, where possible, extending the transferability of the proposed approach across different modalities and application domains.

\newpage
{
    \small
    \bibliographystyle{ieeenat_fullname}
    \bibliography{main}
}

\clearpage
\setcounter{page}{1}
\setcounter{section}{0}
\maketitlesupplementary

\section*{Extended Discussion}
\label{sec:ext_discussion}
While this preliminary study demonstrates strong potential for leveraging semantic representations from vision foundation models in long-tail recognition, several aspects merit deeper investigation to fully realize this approach.

\textbf{Domain Transferability.} Our empirical success raises fundamental questions about when and why latent-space augmentation succeeds. A systematic investigation of vision foundation model embeddings --- particularly quantifying inter-class overlap, analyzing clustering quality across semantic categories, and establishing relationships between pre-training diversity and downstream performance would provide crucial insights. This analysis could explain our contrasting results on the Places-LT benchmark, where scene categories likely exhibit greater semantic ambiguity than object-centric datasets. Understanding these distributional properties would enable principled sampling strategies. Such theoretical grounding would help practitioners predict which visual recognition tasks benefit most from embedding-space augmentation versus traditional fine-tuning approaches.

\textbf{Trade-offs of the Frozen Encoder Paradigm.} Our approach achieves efficiency through single-pass feature extraction and a lightweight linear classifier, yet the quality of synthetic samples remains bounded by the pre-trained backbone's discriminative capacity, as shown in Table~\ref{tab:encoder_ablation}. Class ambiguities or suboptimal decision boundaries in these representations create an irreversible bottleneck, which is particularly problematic for fine-grained datasets where subtle inter-class distinctions matter. For instance, distinguishing between similar bird species or architectural styles requires feature refinements that frozen general-purpose pooled embedding cannot directly provide. This limitation suggests a promising direction: leveraging full spatial feature maps rather than single pooled embeddings. Patch-wise representations could capture fine-grained distinctions while maintaining computational efficiency, potentially through attention-weighted pooling or learnable aggregation mechanisms that still require minimal parameters.

\end{document}